\documentclass[runningheads]{llncs}

\usepackage{eccv}

\usepackage{eccvabbrv}

\usepackage{graphicx}
\usepackage{enumitem}
\usepackage{subcaption} 
\usepackage{booktabs}

\usepackage{tikz}
\usepackage{pgfplots}
\pgfplotsset{compat=1.18}

\pgfplotsset{
  colormap={magma}{
    rgb255=(0,0,4)
    rgb255=(28,16,68)
    rgb255=(79,18,123)
    rgb255=(129,37,129)
    rgb255=(181,54,122)
    rgb255=(229,80,100)
    rgb255=(251,135,97)
    rgb255=(254,194,135)
    rgb255=(252,253,191)
  }
}

\usepackage[accsupp]{axessibility}  

\usetikzlibrary{positioning, shapes.geometric, arrows.meta, calc}

\usepackage{hyperref}

\usepackage{orcidlink}

\begin{document}

\title{From Phase to Phenomenon: Self-Supervised Learning of Subsurface Scattering with Minimal Phase-shift Inputs\thanks{Accepted to ECCV 2026; the final authenticated version will be available via Springer.}}

\titlerunning{From Phase to Phenomenon}

\author{Arjun Majumdar\orcidlink{0009-0009-2321-7011} \and
Raphael Braun\orcidlink{0009-0003-7307-8773} \and
Andreas Engelhardt\orcidlink{0000-0003-1313-3665} \and
Hendrik PA. Lensch\orcidlink{0000-0003-3616-8668}
}

\authorrunning{Majumdar et al.}

\institute{Eberhard Karls Universität Tübingen, Germany
\email{\{arjun.majumdar,raphael.braun,andreas.engelhardt, hendrik.lensch\}@uni-tuebingen.de}}

\maketitle

\begin{abstract}
    We propose a self-supervised pretraining framework for learning sub-surface scattering (SSS) light transport representations from minimal input. Our method leverages a stereo projector--camera setup that captures only eight high-frequency phase-shift profilometry (PSP) images per view to pretrain an encoder in a multi-view, multi-object setting. We introduce a tailored augmentation strategy for PSP-based SSS data, and show that it significantly outperforms standard ImageNet-style augmentations for SSL pretraining.
    The pretrained encoder learns generalizable SSS representations that transfer effectively to downstream tasks, including spatially varying relighting and representation evaluation using a kNN classifier.  Combined with a decoder, the model reconstructs dense scattering footprint responses, trained using a dedicated cost function that improves accuracy, particularly for anisotropic footprints. An overview of our method is presented in Fig.~\ref{fig:method_overview}. Despite using only eight input images per view, our approach generalizes to unseen objects with complex geometry and material properties, achieving high-fidelity reconstructions while requiring orders of magnitude fewer images than prior methods. Our code is publicly available at \href{https://github.com/arjun-majumdar/from-phase-to-phenomenon/tree/main}{GitHub}.

\end{abstract}

\section{Introduction}
\label{sec:intro}

Subsurface Scattering (SSS) occurs when light penetrates the surface of a
material, scatters internally, and exits elsewhere. The scattering depth varies
with the material and its geometry, effectively smoothing the incident light
patterns. Analytical diffusion models describe this process and provide exact
solutions for simple geometries \cite{practical_sss_light_transport}. Recent
deep learning methods extend these solutions to complex surfaces while
preserving efficiency \cite{learned_sss_vicini}. Alternatively, volumetric path
tracing can accurately simulate SSS but requires knowledge about the internal
scattering properties of the material, which are difficult to obtain
\cite{manuka,renderman}. Instead, we adopt an image-based acquisition and
rendering approach that captures each light interaction as an image footprint.
Relighting is then achieved by combining these images with appropriate scaling.
However, capturing the necessary dataset is expensive due to the non-local
nature of SSS. We propose a novel, fully data-driven framework for image-based
modeling of sub-surface scattering (SSS) that learns to accurately predict
spatially varying SSS footprint responses at each surface point in a multi-view,
multi-object setting. Once trained, the model enables scene relighting without
requiring any additional acquisitions. Our method uses a non-contrastive
self-supervised SimSiam objective \cite{simsiam_paper} for pretraining an
encoder backbone, which does not require explicit SSS footprint ground truth
supervision. After pretraining, the learned representations are used to solve
two downstream tasks: {\bf Footprint reconstruction} A decoder is trained to
predict dense SSS footprint responses across multiple views and across diverse
objects. {\bf kNN accuracy} The pretrained encoder learns generalizable SSS
representations, whose quality is evaluated using a zero shot kNN classifier on the
learned embeddings \cite{unsupervised_feature_learning_instance}. This setup
demonstrates that the proposed self-supervised pretraining captures physically
meaningful and generalizable SSS features that transfer effectively across tasks
and objects. 

The encoder is pre-trained in an SSL manner by using only 8 camera captured high-frequency phase shift profilometry (PSP) images as input for each object. PSP has been used for quite some time to capture accurate 3D geometry of scanned objects \cite{tpu_review_paper}, \cite{tpu_comparative_study}. Our encoder is trained on the same sinusoidal patterns that are used for projector-camera registration. Our setup generalizes to unseen views and objects with varying geometries and material properties. We conduct extensive qualitative and quantitative evaluations of the learned footprint responses by rendering objects under a black-white checkerboard virtual illumination pattern and compare the results against camera-captured ground truth. Since our approach focuses on learning and estimating each surface pixel footprint response, we emphasize detailed evaluations of the relit results.

\noindent Our contributions are as follows:
\begin{itemize}
    \item We propose a data-efficient self-supervised framework for learning
    transferable subsurface scattering (SSS) representations from a small
    number of phase-shift profilometry (PSP) images per view. By pre-training
    the encoder, our method significantly reduces acquisition requirements
    compared to prior supervised approaches and enables effective reuse for
    downstream tasks.

    \item We present a physics-aware data augmentation pipeline specifically
    designed for high-frequency structured-light PSP inputs, enabling stable and
    effective self-supervised pretraining on SSS data.

    \item We design a task-specific loss formulation that accounts for the
    highly skewed and anisotropic distribution of SSS footprint responses. This
    tailored objective stabilizes training and improves reconstruction accuracy
    by explicitly encouraging the learning of anisotropic scattering behavior.

    \item We train a lightweight decoder to
    directly predict anisotropic pixel-level SSS footprint responses from
    encoded representations of projected high-frequency PSP patterns with explicit SSS footprint
    supervision. The encoder's learned representations are evaluated using a standard kNN
    classifier \cite{unsupervised_feature_learning_instance}.

    \item We demonstrate strong generalization across unseen objects, geometries, materials, and viewpoints, validated through extensive qualitative and quantitative experiments.
\end{itemize}

\begin{figure}
    \centering
    \input{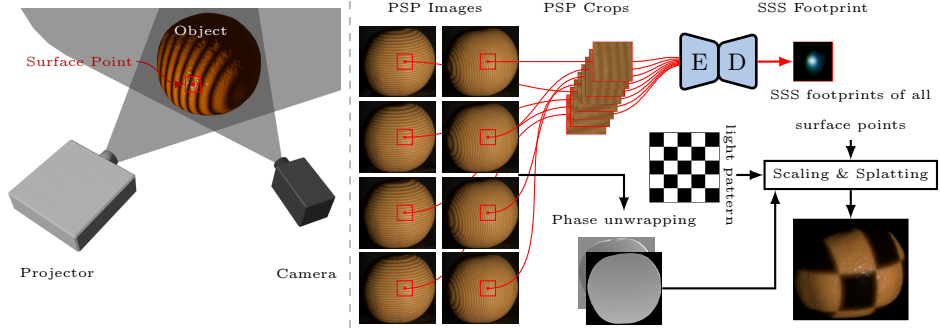}
    \caption{\textbf{Left:} Our method works on an uncalibrated stereo projector-camera setup. \textbf{Center:} We capture images of the object under eight high-frequency sine wave patterns (PSP Images). Registration of camera- to projector-pixels is achieved via phase unwrapping. Our method estimates the SSS point spread function for any given surface point from a local crop of the PSP images. Relighting is achieved by scaling those SSS footprints for every surface point with the light they would receive from the desired light pattern, and splatting them onto a target canvas.}
    \label{fig:pipeline}
\end{figure}

\section{Background}
\label{sec:background}

Acquisition of material properties has a long history, including measurement of SSS parameters and general relightable representations such as reflectance fields~\cite{debevec2000acquiring}. Early methods directly sampled the impulse response to incident illumination, while later work employed fixed wavelet noise patterns~\cite{peers2003wavelet}, compressive sensing~\cite{peers2009compressive}, and adaptive acquisition strategies~\cite{sen2005dual,garg2006symmetric,o2010optical,o2012primal} that exploit the directional duality of light transport to accelerate the capture of general light transport matrices. Despite these advances, the number of required acquisition patterns remains large. In contrast, our method requires only eight input images per new object and view after an initial training stage across multiple objects to recover spatially varying pixel footprints for SSS.

Following the introduction of the first practical BSSRDF model~\cite{jensen2001practical}, SSS parameters have been estimated using dedicated point-wise illumination setups~\cite{jensen2001practical,weyrich2006analysis}.
For complete objects, Goesele et al.~\cite{goesele2004disco} presented a system in which a single laser beam sequentially illuminates each surface point to measure the resulting footprint at the camera, requiring millions of individual measurements.

Recent neural relighting methods also target objects exhibiting SSS~\cite{zheng2021neural,lyu2022neural,yu2023learning,zhu2023neural,dihlmannsubsurface}, but they typically assume distant illumination instead of incident light patterns and rely on one-light-at-a-time (OLAT) data for training.

Structured-light techniques, particularly those based on high-frequency patterns, have been used to separate reflected radiance components into local and global illumination effects~\cite{nayar2006fast}. Several 3D scanning approaches employ high-frequency PSP patterns for objects with SSS~\cite{fuchs2008combining,mod_phase_shift_3d,polarization_phaseshift,micro_phaseshift,embed_phaseshift}. However, these methods primarily aim to suppress subsurface scattering artifacts to improve depth estimation rather than to measure or represent the scattering itself for relighting. Geiger et al.~\cite{improved_topo_sss} also use structured light projection to correct topographic errors caused by volumetric light transport inside scattering materials, analyzing such effects through Monte Carlo simulations and introducing correction techniques based on quantified light propagation. Other studies~\cite{kienle1996spatially,kikuchi2023development} leverage high-frequency binary patterns and polarization filters to directly measure local isotropic scattering parameters of human skin.

Vicini et al.~\cite{learned_sss_vicini} proposed a neural framework for efficiently sampling Bidirectional Surface Scattering Reflectance Distribution Functions (BSSRDFs) on complex 3D surfaces. Their system combines three components for conditioning a Conditional-VAE and MLP architecture. Majumdar et al. \cite{ourpaper} present a U-Net \cite{unet_paper} based method which leverages 3D scanning and a stereo projector-camera setup to learn pixel-level SSS responses for accurate, high-resolution relighting and generalization across materials and views. They use no data augmentation and/or SSL method and feed in raw inputs directly as inputs to map to SSS footprint responses for all surface points requiring expensive data acquisition and consequent processing. They train the encoder-decoder in an end-to-end manner. Therefore, their encoder cannot be used in a standalone manner for different tasks.  Our method does not suffer from these shortcomings. To the best of our knowledge, we are the first to introduce and use data augmentation specifically tailored for high-frequency PSP SSS input patches for SSL non-contrastive pre-training. The pre-trained encoder is then used for the downstream task of reconstructing spatially varying, potentially anisotropic scattering footprints. Pre-training the encoder results in a simpler and more efficient training pipeline. The resulting neural model produces accurate, relightable representations of scattering behavior. 

\subsection{Phase-shifted Profilometry}
Our acquisition of subsurface scattering representations builds on N-step Phase-Shifting Profilometry (PSP), a technique commonly used for high-accuracy 3D scanning ~\cite{tpu_comparative_study,psp_review_paper,mod_phase_shift_3d,polarization_phaseshift,micro_phaseshift,embed_phaseshift}. In a calibrated camera--projector setup, a sequence of $N$ phase-shifted sinusoidal fringe patterns is projected onto the object and captured by the camera. Each projected pattern can be mathematically defined as
\begin{equation}
I^p(x^p, y^p) = a^p + b^p \cos(2\pi f_0^p x^p + \delta_n),
\end{equation}
where $(x^p, y^p)$ are projector coordinates, $a^p$ is the average intensity, $b^p$ the amplitude, $f_0^p$ the fringe frequency, and $\delta_n = 2\pi n / N$ the phase shift.

The corresponding camera captured images are modeled as
\begin{equation}
I_n(x, y) = A(x, y) + B(x, y) \cos \left(\phi(x, y) - \frac{2\pi n}{N}\right),
\end{equation}
where $A(x,y)$ denotes background illumination, or indirectly scattered illumination, $B(x,y)$ the modulation (linked to reflectivity), and $\phi(x,y)$ the wrapped phase, computed as
\begin{equation}
\phi(x, y) = \tan^{-1} \frac{\sum_{n=0}^{N-1} I_n(x, y) \sin(2\pi n/N)}{\sum_{n=0}^{N-1} I_n(x, y) \cos(2\pi n/N)}.
\end{equation}
The wrapped phase $\phi(x,y) \in [-\pi, \pi]$ is converted to an unwrapped phase $\Phi(x,y) = \phi(x,y) + 2\pi k(x,y)$, where $k(x,y)$ indicates fringe order, correlating directly with scene depth in stereo geometry.

Phase unwrapping, i.e.\ determining global offsets based on locally recovered phases $\phi(x, y)$, can be performed spatially using neighboring pixels or temporally using phase shifts over time. Temporal methods are generally more robust, especially near surface discontinuities~\cite{tpu_comparative_study}. However, for translucent or highly scattering materials, traditional PSP often fails because subsurface scattering distorts the low-frequency phase maps required for reliable unwrapping~\cite{mod_phase_shift_3d,polarization_phaseshift}. We capture eight images for $N=4$ step PSP: four with height-varying and four with width-varying sinusoidal patterns. Example unwrapped phase maps for the orange is depicted in Fig.~\ref{fig:pipeline}. 

\begin{figure}[t]
\centering

\begin{subfigure}[t]{0.3\linewidth}
    \centering
    \includegraphics[width=\linewidth]{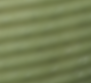}
    \caption{Apple}
\end{subfigure}
\hfill
\begin{subfigure}[t]{0.3\linewidth}
    \centering
    \includegraphics[width=\linewidth]{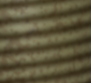}
    \caption{Pear}
\end{subfigure}
\hfill
\begin{subfigure}[t]{0.3\linewidth}
    \centering
    \includegraphics[width=\linewidth]{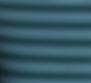}
    \caption{Star}
\end{subfigure}

\caption{\textbf{Effect of SSS on PSP images.} Strongly scattering materials (apple) blur projected patterns and reduce contrast due to wider scattering footprints, whereas weakly scattering objects (pear, star) preserve sharper sinusoidal structures.}
\label{fig:psp_show_scattering}

\end{figure}

SSS affects captured PSP images by spatially spreading the reflected light beneath the surface. Materials with stronger scattering blur the projected sinusoidal patterns and reduce their local contrast, which will be encoded into the latent representations. For example, highly scattering objects such as apples exhibit significantly attenuated and smoother PSP patterns, whereas objects with weaker scattering, such as pears, preserve sharper pattern contrasts which is shown in Fig. \ref{fig:psp_show_scattering}.

\section{Method}

\begin{figure}[htb]
    \centering
    \resizebox{0.9\linewidth}{!}{\input{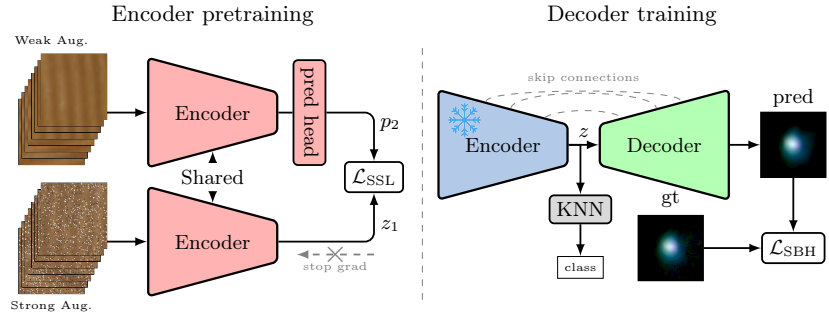}}
    \caption{\textbf{Overview of proposed framework.} \textbf{Left:} SSL
    pretraining of shared encoder using 2 augmented PSP patches, minimizing
    $L_{\text{SSL}}$. \textbf{Right:} The decoder is trained to predict SSS
    footprints in a supervised setting, exploiting the embedding space of the
    frozen encoder. It is optimized based on ground truth SSS footprints and an
    auxiliary classification task. Classification is done with a zero-shot kNN
    approach directly on the encoder features.}
    \label{fig:method_overview}
\end{figure}

We present an overview of our method in Fig.~\ref{fig:method_overview} which on LHS shows SSL pre-training of the encoder, while RHS shows the frozen encoder used for the two downstream tasks, viz., training the decoder on SSS footprint responses and kNN zero-shot classification.

\subsection{Self-Supervised Learning}
Self-Supervised Learning (SSL) in computer vision aims to learn transferable and generalizable feature representations without relying on manual annotations. Most modern SSL approaches are based on similarity learning, where a backbone network $f$ is trained to map semantically related inputs to nearby points in a high-dimensional embedding space. Typically, these related inputs are simulated during training as different augmentations of the same image.

Many contemporary methods adopt a Siamese architecture consisting of weight-sharing neural network branches that process paired views of an image. Although this framework enables effective representation learning, it also introduces the risk of \emph{representation collapse}, where the network converges to a degenerate solution by mapping all inputs to an almost constant embedding. 

Extensive prior work has analyzed this collapse phenomenon and proposed mechanisms to prevent it. In general, SSL methods can be categorized into three groups: {\bf Contrastive methods:} These approaches explicitly pull positive (similar) embeddings closer while pushing negative (dissimilar) embeddings apart. The use of negative samples prevents trivial constant mappings and stabilizes training \cite{simclr_paper}, \cite{moco_paper}, \cite{swav_paper}. {\bf Non-contrastive methods:} These methods align positive pairs without relying on negative samples. To avoid collapse, architectural asymmetries (e.g., predictor networks, stop-gradient operations) or implicit constraints are introduced \cite{byol_paper}, \cite{simsiam_paper}. {\bf Regularization-based methods:} These approaches impose explicit statistical constraints on the embedding space, such as encouraging feature variance, decorrelation across embedding dimensions, or enforcing the cross-correlation matrix to approximate identity \cite{barlowtwins_paper}, \cite{vicreg_paper}, \cite{whitening_ssl}. Such constraints ensure diverse and non-degenerate representations. This taxonomy provides a unified perspective on how modern SSL frameworks prevent collapse while learning meaningful visual representations.

We propose a self-supervised pretraining framework for learning sub-surface
scattering (SSS) light transport representations from high-frequency phase-shift
profilometry (PSP) images. We pretrain an encoder in a multi-view, multi-object
setting to learn a generalizable SSS representation. We then train a decoder to predict the SSS responses and a prediction head to estimate the material class from those representations in a supervised manner. Figure~\ref{fig:method_overview} visualizes our training pipeline for both stages.

\subsection{Data Acquisition}
For the encoder, we capture 8 PSP images for every object. For training the
decoder, we further capture the ground truth scattering footprints by illuminating individual dots on the surface for several training objects. 
We use a stereo projector-camera setup to capture both pixel-level SSS responses (point spread functions PSF) and the PSP images as shown in Fig.~\ref{fig:pipeline}. The acquisition setup and basic image processing
follow~\cite {ourpaper}.
The processing pipeline for all captured images includes the acquisition of raw
Bayer images, median filtering each RGGB channel with a $3\times 3$ filter,
demosaicing with Malvar's algorithm ~\cite{malvar_demosaicing} and finally
masking the objects using SAM~\cite{sam_paper}. Further details are in supplemental (S1).

For the encoder, we use high-frequency 
PSP patterns with 4 shifts. This means that we only capture 8 images per-view per
object, 4 with vertical and 4 with horizontal orientation. 
No more than these 8 images per view are are necessary to train the encoder and to do the final inference of the scattering response footprints. We crop patches of size $(90 \times 90)$ pixels from those 8 images and provide them as input to the encoder as indicated in Fig.~\ref{fig:pipeline} and Fig.~\ref{fig:method_overview}.

\begin{figure}[htb]
    \centering
    \input{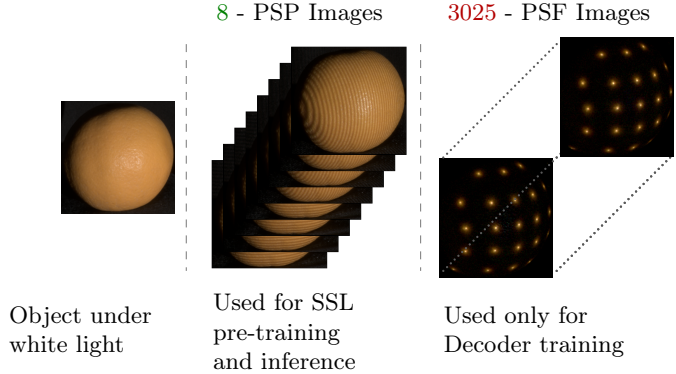}
    \caption{For pre-training the Encoder with SSL we only use 8 PSP images per
    object. The point spread
    functions (PSF) of every surface point are captured in parallel with a
    spacing of 55 pixels, which requires 3025 images. Those images are only
    captured and used to train the Decoder once. The decoder generalizes to
    unseen objects during inference. Thus for relighting new objects we only
    have to capture 8 PSP images.
    }
    \label{fig:psp_psf}
\end{figure}

\textbf{Local SSS footprint responses} are captured by projecting dot patterns that
illuminate individual points on the object's surface. This grid pattern enables the parallel acquisition of SSS
responses for multiple surface points in one image. The distance between each
illuminated pixel in projector space is chosen such that the footprints do not
overlap (55 pixels in both $x$ and $y$ directions),
The result of this acquisition can be seen on the right side of Figure~\ref{fig:psp_psf}. The grid
pattern is shifted that every surface point is illuminated once. This
acquisition is repeated for multiple views per object and for multiple different
{\it training} objects. 
The total number of such SSS footprint response samples = 1126827 in
total for 5 objects (Green apple, Orange, Pear, Star and Shovel) with up to four
views each.
For inference on novel objects, the trained
encoder-decoder network can directly predict the footprints purely based on the 8 PSP
images, not requiring any captured footprints. 

\subsection{SSL Pretraining}
Typical SSL methods use either ResNet-50 or ViT backbones \cite{simclr_paper,simclrv2_paper,moco_paper,vicreg_paper,byol_paper,msbyol_paper}. We do not adopt these architectures since our framework additionally requires a decoder to learn SSS footprint responses. A vanilla U-Net block $[\text{conv} \rightarrow \text{BN} \rightarrow \text{ReLU}] \times 2$ is less effective than residual blocks for learning stable embeddings and is more prone to dimensional collapse. Therefore, we design a custom architecture inspired by pre-activation residual blocks \cite{identity_mappings_preactivation_resnet}. Detailed encoder and decoder architectures are provided in supplemental (S2). We pre-train the encoder with SimSiam's~\cite{simsiam_paper} non-contrastive method to learn generic representations of the high-frequency PSP input patches for each point on the object's surface. We provide ablations of using SimSiam's hyper-parameters vs. our modified ones together with dimensional collapse plots in the supplemental. During training, the sets of PSP patches are randomly augmented to two different views (Sec.~\ref{sec:augmentation}). The training loss forces the output latent vectors to be invariant to the augmentations (see Fig.~\ref{fig:method_overview}).

The crops for pretraining are extracted from every point on the object's surface within the SAM mask. For each point, a $(90,90)$ pixel patch around the pixel is cropped out for all 8 PSP images. They are used as input for minimizing the negative cosine similarity on normalized vectors for the SimSiam method \cite{simsiam_paper}, which can be mathematically written as:
\begin{equation}
    \mathcal{\mathcal{L}_{\text{SSL}}}(p_1, z_2) = - \frac{p_1}{\|p_1\|_2} \cdot \frac{z_2}{\|z_2\|_2}.    
\end{equation}

\subsection{Augmentation}\label{sec:augmentation}
One of the most important factors in SSL pre-training is the employed data
augmentation. There are standard data augmentation pipelines that
are tailored for natural images. We have very specific types of images, namely
PSP SSS patches, for which we developed our own specialized data augmentation
pipeline.

\begin{figure}[t]
\centering
\setlength{\tabcolsep}{2pt}

\begin{tabular}{cccccc}
\includegraphics[width=0.15\linewidth]{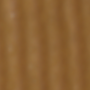} &
\includegraphics[width=0.15\linewidth]{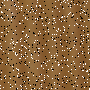} &
\includegraphics[width=0.15\linewidth]{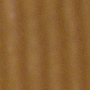} &
\includegraphics[width=0.15\linewidth]{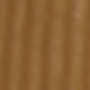} &
\includegraphics[width=0.15\linewidth]{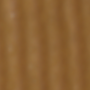} &
\includegraphics[width=0.15\linewidth]{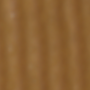} \\
 Orig. & Salt\&Pepper & ph. jitter & flip & shuffle & blur
\end{tabular}

\caption{Demonstration of our PSP specific strong augmentations for one input patch.}
\label{fig:augmentations}

\end{figure}

Following \cite{msbyol_paper}, instead of strong-strong augmentations on both network branches,
we apply a weak-strong augmentation pipeline
to stabilize training and to obtain better generalizing representations. 
Consequently, we formulate our custom data augmentations for high-frequency PSP input patches as:
{\bf Weak augmentation:} We follow \cite{msbyol_paper} with random-resized cropping and slight color jitter.
{\bf Our Strong augmentation:} salt\&pepper noise, Gaussian blur,
horizontal or vertical flips on all  PSP images, 
random intensity scaling, random phase-jitter (add Gaussian
    (phase-like) noise to RGB images) and randomly shuffled PSP images (to break
    fixed $\frac{2\pi}{N}$ phase-shifts symmetry between the camera
    images).
The effect of our new strong augmentations on a patch is visualized in Fig.~\ref{fig:augmentations}. 

\noindent In principle, the trained representations $z$ could be useful for any downstream task. We specifically use them to train a robust footprint prediction
decoder, which generalizes to unseen objects. 

\subsection{Supervised Decoder Training}
To estimate the SSS footprints from PSP patches, the encoder is kept frozen while the decoder is trained in a supervised fashion, as shown on the right side of Fig.~\ref{fig:method_overview}. For each pixel, we crop a $90\times 90$ patch from the PSF images centered at that pixel and encode it into a representation $z$. The decoder then predicts the local SSS response, supervised by camera-captured ground truth SSS responses for surface points on the object using our Self-Balanced Hybrid Loss $\mathcal{L}_{\text{SBH}}$ (Sec.~\ref{sec:decoder_loss}). 

In the decoder, each decoder stage performs progressive spatial upsampling with attention-gated skip fusion using sub-pixel upsampling \cite{pxshuffle_paper}.
This formulation can be interpreted as a \emph{machine translation} problem, where the network maps PSP patch observations to a different output domain corresponding to local SSS footprint responses. Training the decoder object-by-object leads to catastrophic forgetting; therefore, we randomly sample patches across all objects and views during training.

\subsubsection{Relighting Pipeline}
Let $L$ denote the virtual projector image and $\pi$ the camera--projector pixel correspondence recovered from the unwrapped phase. For each surface pixel $p$: (i) encode PSP crop $90{\times}90$ centered on $p$ and decode to obtain predicted SSS footprint $\mathbf{PSF}_{p}\in\mathbb{R}^{90\times90\times3}$; (ii) look up projector color $c(p)=L(\pi(p))$ $\rightarrow$ this is the \textbf{scaling} step; (iii) \textbf{splat} each scaled footprint into relit canvas and accumulate: $I(q)\;=\;\sum_{p}\mathbf{PSF}_{p}(q-p)\cdot c(p)$, where $c(p)$ scales the contribution of light entering at $p$ and the sum over $p$ performs the splat-and-accumulate.

\subsection{Self-Balanced Hybrid Loss}\label{sec:decoder_loss}
The light footprints caused by subsurface scattering typically have a strong peak in the point that is illuminated with an exponential radial falloff. We designed a physics-aware hybrid loss tailored for reconstructing those footprints. Our loss consists of five components:  
\begin{align*}
\mathcal{L}_{\text{SBH}} = \lambda_{1}\mathcal{L}_{1} + \lambda_{\log}\mathcal{L}_{\log} + \lambda_{\mathrm{SSIM}}\mathcal{L}_{\mathrm{SSIM}} + \lambda_{\nabla}\mathcal{L}_{\nabla} + \lambda_{\mathrm{wMSE}}\mathcal{L}_{\mathrm{wMSE}}.
\end{align*}
\begin{description}
    \item[$\mathcal{L}_1$]: Standard L1 loss as data term. Works well on low-intensity data and prevents over-smoothing. 
    \item[$\mathcal{L}_{\log}$]: L1 loss on logarithmic data. This increases sensitivity to errors in faint regions.
    \item[$\mathcal{L}_{\mathrm{SSIM}}$]: Structural similarity loss as perceptual component.
    \item[$\mathcal{L}_{\nabla}$]: L1 loss on channel normalized gradients. The normalization of the gradient of each channel prevents color-scale collapse.
    \item[$\mathcal{L}_{\mathrm{wMSE}}$]: Pixel-wise variance across the batch is used to weight the loss, placing greater emphasis on high-variance regions of the SSS footprint by using the relative variance as weight:
            $\sigma^2_{cij} = \operatorname{Var}_{b}(Y^{(b)}_{cij})$ and
            $w_{cij} = 1 + \sigma^2_{cij} / (\bar{\sigma}^2 + \varepsilon)$.
            We use this pixel wise weight $w_{cij}$ in a standard MSE loss.
\end{description}

\noindent The proposed objective jointly sharpens the peaks ($\mathcal{L}_{\nabla}$,
$\mathcal{L}_{\mathrm{wMSE}}$), preserves SSS tails ($\mathcal{L}_{\log}$),
stabilizes global intensity ($\mathcal{L}_{1}$), and maintains perceptual
structure ($\mathcal{L}_{\mathrm{SSIM}}$). Batch-adaptive variance weighting enables self-balancing behavior, mitigating the intrinsic smoothing bias of convolutional networks. In our experiments, we set the loss weights to 
$\lambda_1=1,\ \lambda_{\mathrm{WMSE}}=5,\ \lambda_{\nabla}=10,\ \lambda_{\mathrm{SSIM}}=0.5,\ \lambda_{\log}=1$.

\subsection{Relighting}
Given our trained encoder and decoder, we can relight any object given eight
sinusoidal fringe images. For every surface point, we
crop $90\times90$ patches around the point from the PSP images, obtain their latent representation with the encoder and
finally decode the latent to obtain the pixel's SSS response footprint.
Relighting then boils down to
scaling those responses with the light received from the virtual projector image
and splatting the scaled footprint onto the image canvas. This is visualized on the right side of Fig.~\ref{fig:pipeline}.

\section{Results}

\subsection{Training Details}
When using SimSiam's hyperparameters, we observe a dimensional collapse due to the different data distribution. Therefore, we modify the parameters and start with an initial lr=0.05 \textit{without scaling} accompanied with lr warmup (20 warmup epochs), followed by cosine decay without warm restarts. This modification allows us to get a stable training without dimensional collapse. Training details, loss curve visualizations, dimensional collapse and further analysis are in the supplemental.

\begin{figure}[t]
\centering
\setlength{\tabcolsep}{2pt}
\begin{tabular}{@{}c@{\hspace{2pt}}c@{\hspace{2pt}}c@{}}

\includegraphics[width=0.3\linewidth,height=0.3\linewidth]{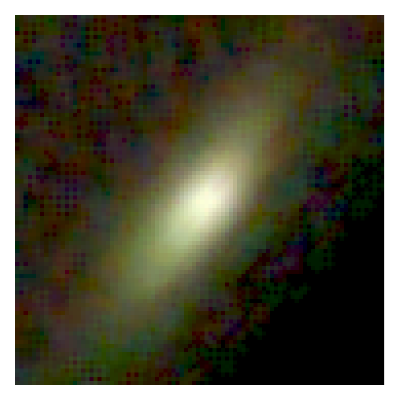} &
\includegraphics[width=0.3\linewidth,height=0.3\linewidth]{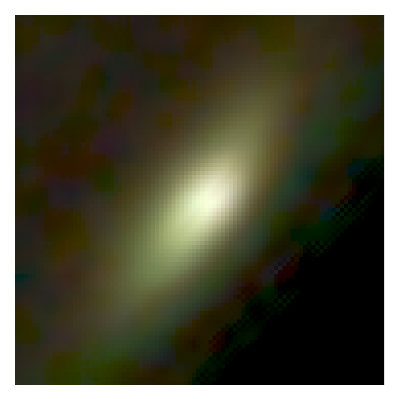} &

\begin{tikzpicture}[baseline=(img.south)]
  \node[anchor=south west,inner sep=0] (img) at (0,0)
    {\includegraphics[width=0.3\linewidth,height=0.3\linewidth]{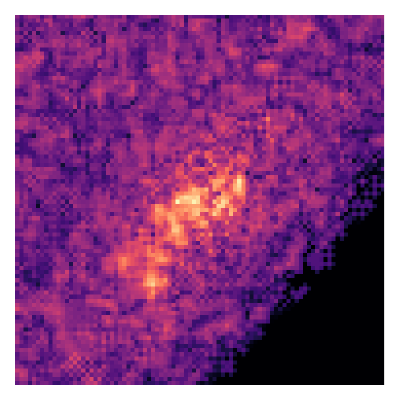}};

 \begin{axis}[
  at={(img.south east)},
  anchor=south west,
  hide axis,
  scale only axis,
  height=0.3\linewidth,
  width=0.045\linewidth,
  xmin=0, xmax=1,
  ymin=0, ymax=1,
  colorbar,
  colormap name=magma,
  point meta min=0,
  point meta max=6.7402665e-3,
  colorbar style={
    width=0.18cm,
    ytick={0,3e-3,6.7e-3},
    yticklabels={0,3e-3,6.7e-3},
    title style={yshift=-2pt},
  }
]
\addplot[draw=none, forget plot] coordinates {(0,0) (1,1)};
\end{axis}
\end{tikzpicture}

\\[-1pt]
Ground Truth & Output & Abs Difference
\end{tabular}

\caption{Qualitative comparison of SSS footprints. Note the denoised quality of the predicted strongly anisotropic footprint.}
\label{fig:psf_images}
\end{figure}

\subsection{Assessment of SSS Footprint Reconstructions}
\begin{figure}[t]
    \centering
    \setlength{\tabcolsep}{2pt}

    \begin{subfigure}[t]{0.48\linewidth}
        \centering
        \includegraphics[width=\linewidth]{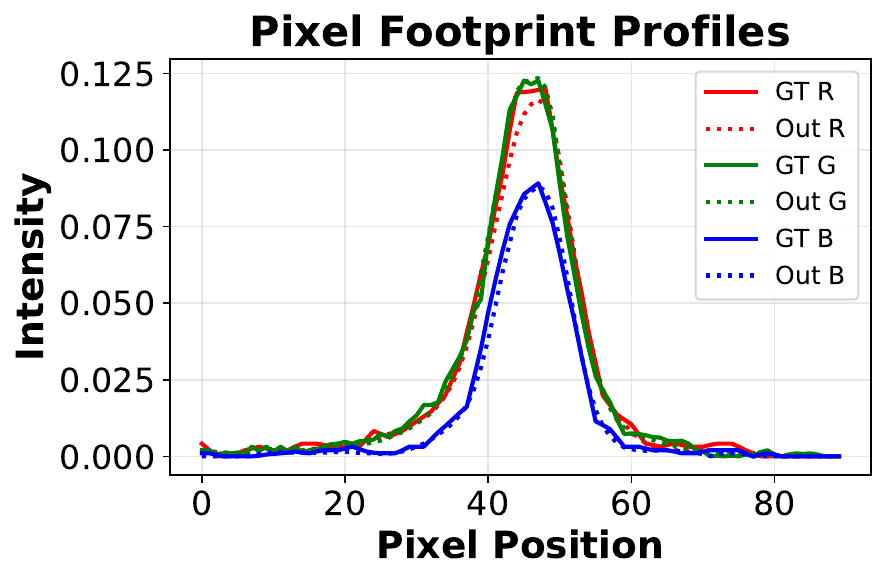}
        \caption{Pixel footprint profiles. Solid lines denote ground truth, dotted lines show the network prediction.}
    \end{subfigure}
    \hfill
    \begin{subfigure}[t]{0.48\linewidth}
        \centering
        \includegraphics[width=\linewidth]{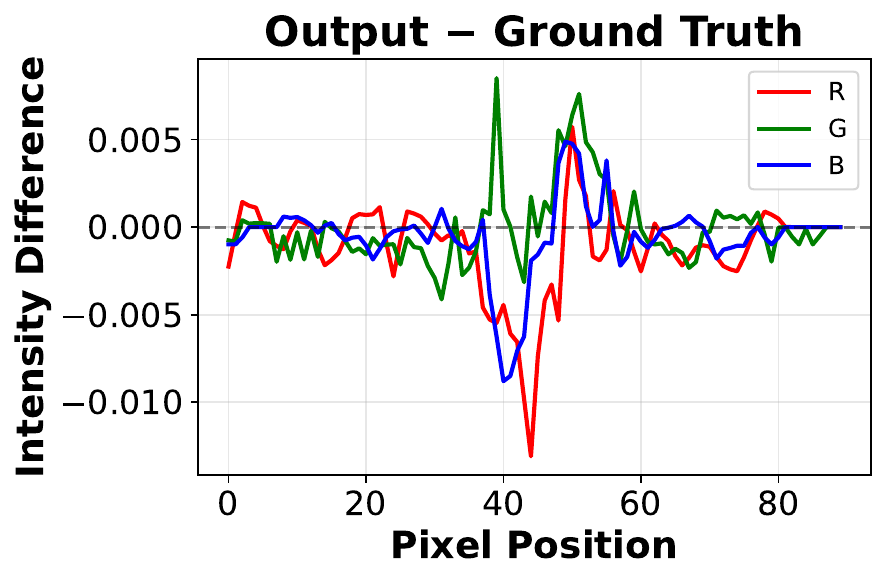}
        \caption{Difference between prediction and ground truth profiles.}
    \end{subfigure}

    \caption{Comparison: predicted vs. ground-truth SSS footprint responses, showing one scan line for a representative patch. Note the faithful prediction from the PSP input even in the tail of the scattering profile. }
    \label{fig:psf_profiles}
\end{figure}

A qualitative comparison of the camera-captured GT footprint responses vs.\ their corresponding output responses, along with L1-difference aggregated for RGB channels, is shown in Fig.\ \ref{fig:psf_images}. Each footprint response patch is $(90,90,3)$ pixels. The trained encoder-decoder network accurately reconstructs the responses even for strongly anisotropic footprints. The apparent camera noise of the GT capture is significantly reduced. To also visualize the quantitative behavior, Fig.\ \ref{fig:psf_profiles} shows the intensity profile across the same footprint as in Fig.\ \ref{fig:psf_images}. The reconstruction accurately follows the ground truth for the three color channels with minor deviations in the peak and precise shape in the extended tail. For more visualizations and reconstructed footprints see the supplemental. Aggregated reconstruction metrics for unseen test objects are reported in Tab.~\ref{tab:test_distribution_quant_metrics}.

\begin{table}[t]
\centering

\begin{minipage}[t]{0.48\linewidth}
\centering
\caption{Aggregated quantitative reconstruction metrics for different test objects and views (MSE$\times10^{-5}$, LPIPS$\times10^{-2}$).}
\label{tab:test_distribution_quant_metrics}
\begin{tabular}{lcccc}
\toprule
\textbf{Object} & \textbf{MSE}$\downarrow$ & \textbf{PSNR}$\uparrow$ & \textbf{SSIM}$\uparrow$ & \textbf{LPIPS}$\downarrow$ \\
\midrule
Apple & 4.1 & 44.67 & 0.96 & 8.4 \\
Orange & 4.9 & 43.84 & 0.94 & 8.6 \\
Crab & 3.5 & 47.98 & 0.98 & 6.5 \\
Hand & 3.8 & 42.51 & 0.96 & 6.5 \\
\bottomrule
\end{tabular}
\end{minipage}
\hspace{0.04\linewidth}
\begin{minipage}[t]{0.42\linewidth}
\centering
\caption{kNN accuracy of PSP-specific augmentations show improved embedding quality compared to ImageNet ones.}
\label{tab:knn_acc_data_aug}
\begin{tabular}{lcc}
\toprule
\textbf{Object} & \textbf{PSP}$\uparrow$ & \textbf{ImageNet}$\uparrow$ \\
\midrule
Apple & \textbf{99.6} & 99.04 \\
Pear & \textbf{99.99} & 98.74 \\
Orange & \textbf{99.99} & 98.92 \\
Star & \textbf{99.95} & 99.93 \\
Shovel & \textbf{91.43} & 84.91 \\
\bottomrule
\end{tabular}
\end{minipage}

\end{table}

\subsection{Relighting Results}
\begin{figure}[htb]
    \centering
    \begin{tabular}{ccccc}
        Cam GT & Relit & Cam GT & Relit\\
        \includegraphics[height=0.13\linewidth]{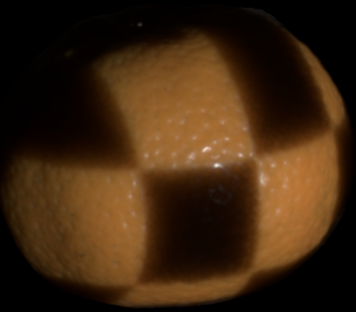} &  
        \includegraphics[height=0.13\linewidth]{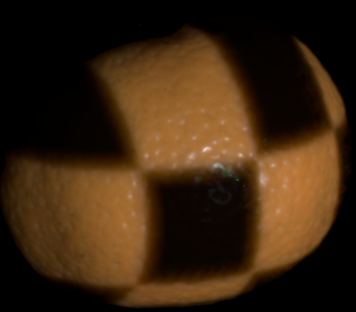}&
        \includegraphics[width=0.13\linewidth, angle=90]{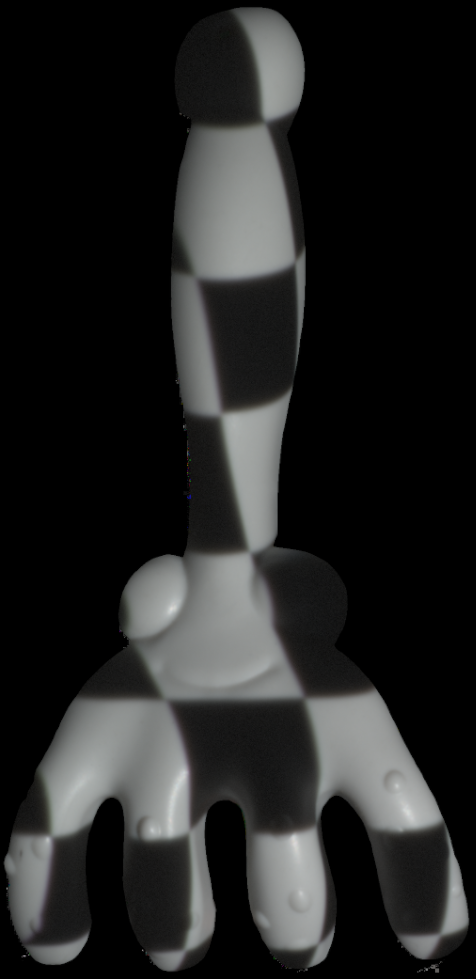}&
        \includegraphics[width=0.13\linewidth, angle=90]{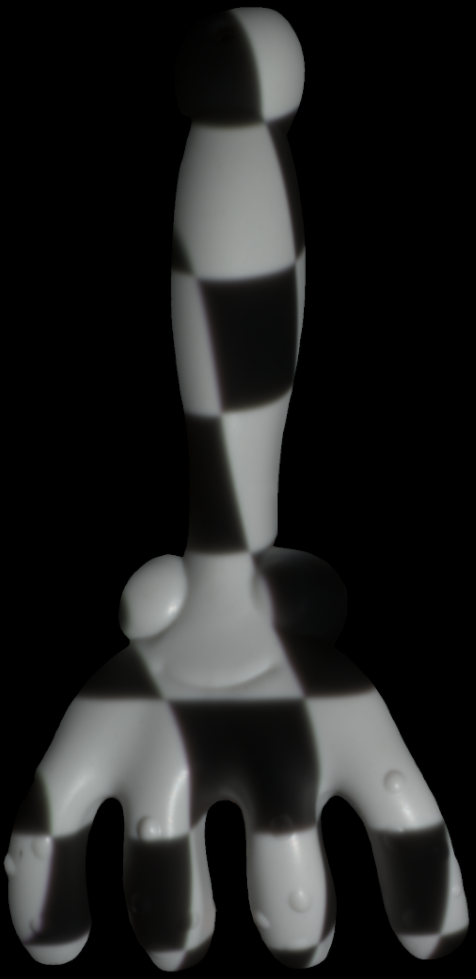}
        \\
        \multicolumn{2}{c}{Tangerine} & \multicolumn{2}{c}{Sand Rake}
    \end{tabular}
    \caption{Qualitative comparisons on two test objects. Cam GT is a real photo of the object lit by the projector, Relit is the result of our relighting. Note: Those objects have never been seen during training of the encoder or decoder. }
    \label{fig:test_frontview_qualitative_comparisons}
\end{figure}

For relighting, we first reconstruct the SSS response footprints for every individual pixel and then can use any virtual projector image matching the original projector's resolution (1920x1080). Qualitative relighting results on unseen objects are shown in Fig.~\ref{fig:test_frontview_qualitative_comparisons}. 
The precise phase unwrapping ensures perfectly aligned incident light patterns. 
The trained network faithfully reconstructs the objects' scattering appearance, in particular, the edges between lit and unlit areas, where SSS leads to smoothed transitions. On the tangerine, even the highlights are properly shown. In the toy sand rake example, the scattering and interreflections between the complex geometry of the fingers are correctly reproduced. 

\begin{figure}[htb]
\centering
\setlength{\tabcolsep}{2pt}

\begin{tabular}{c ccc ccc}

 & \multicolumn{3}{c}{Apple} & \multicolumn{3}{c}{Crab} \\

\rotatebox{90}{Cam GT} &
\includegraphics[width=0.15\linewidth]{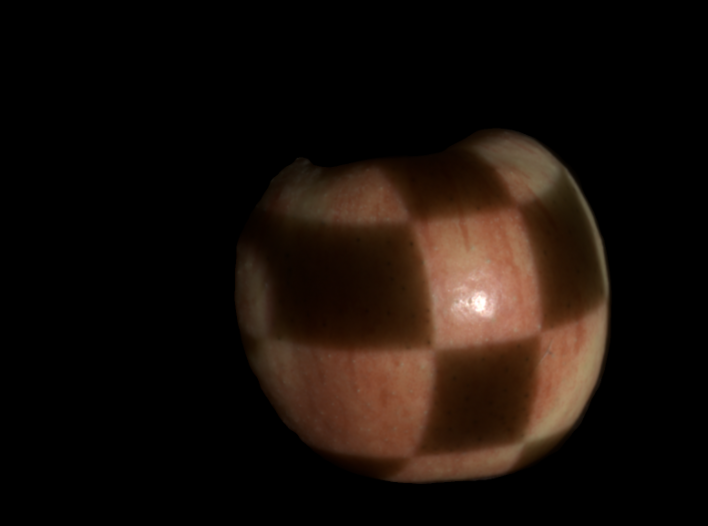} &
\includegraphics[width=0.15\linewidth]{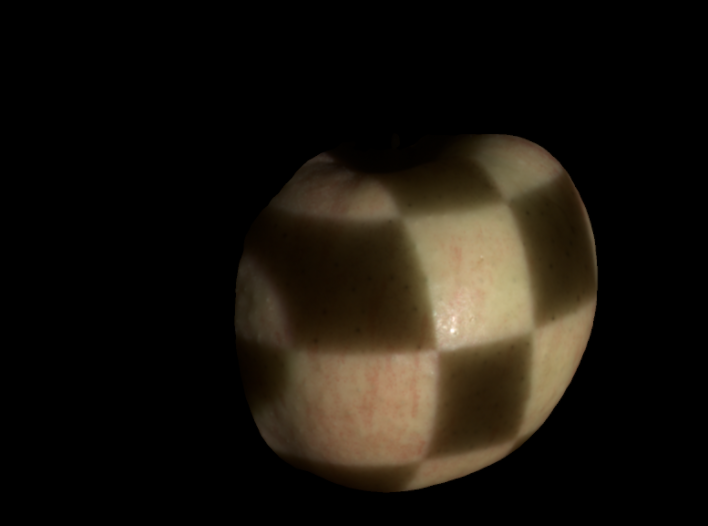} &
\includegraphics[width=0.15\linewidth]{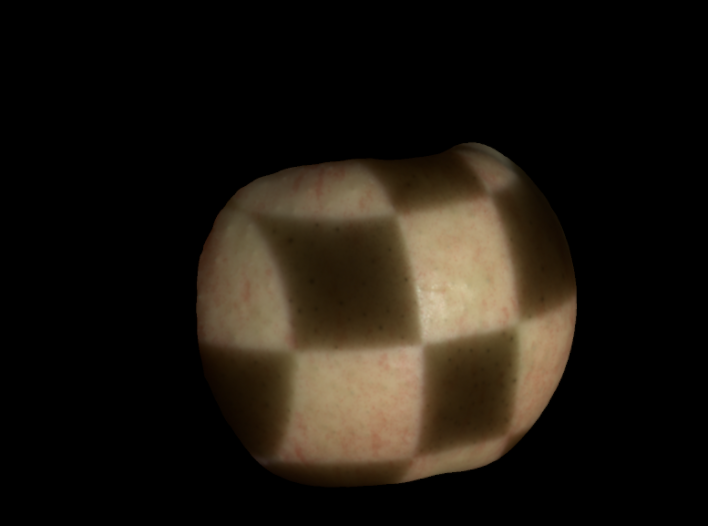} &
\includegraphics[width=0.15\linewidth]{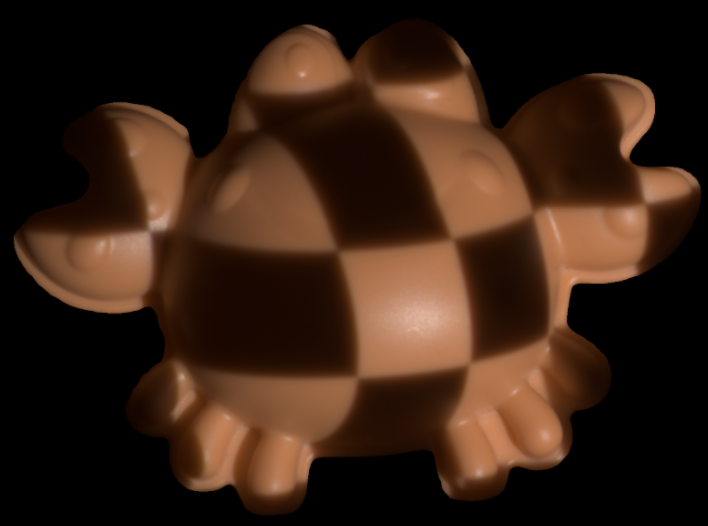} &
\includegraphics[width=0.15\linewidth]{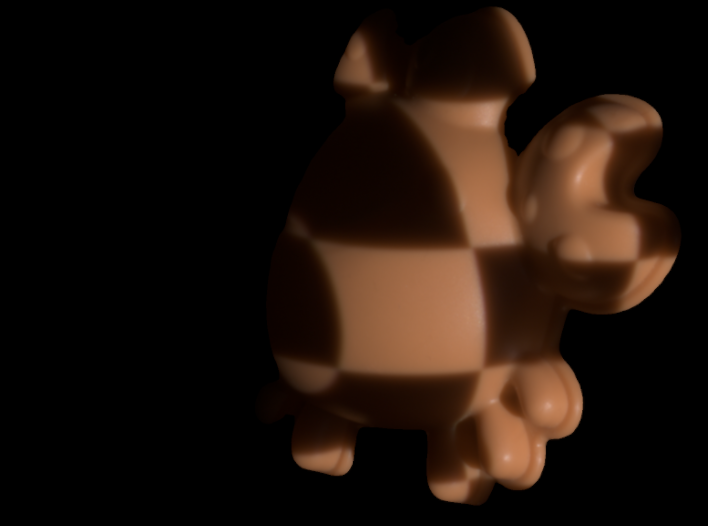} &
\includegraphics[width=0.15\linewidth]{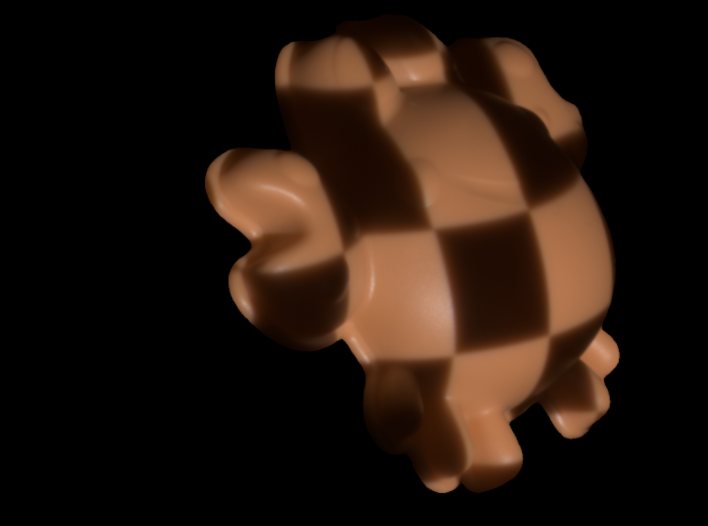} \\

\rotatebox{90}{Relit} &
\includegraphics[width=0.15\linewidth]{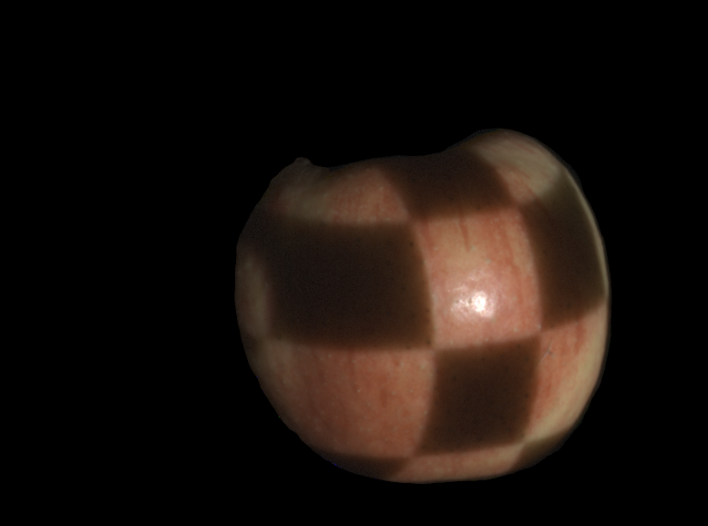} &
\includegraphics[width=0.15\linewidth]{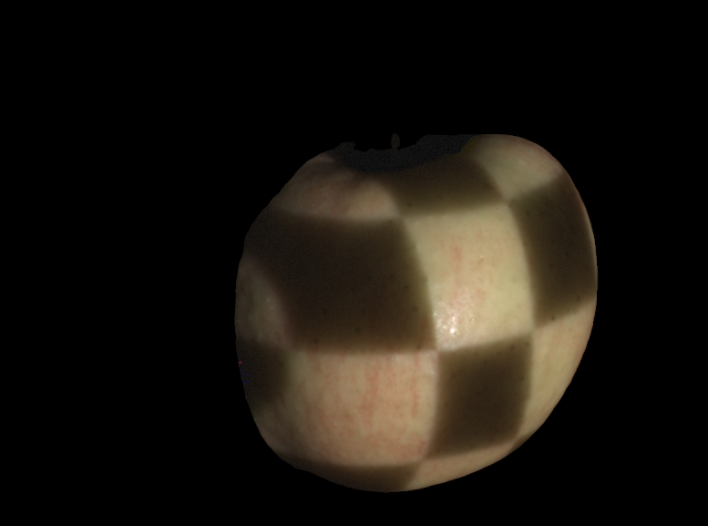} &
\includegraphics[width=0.15\linewidth]{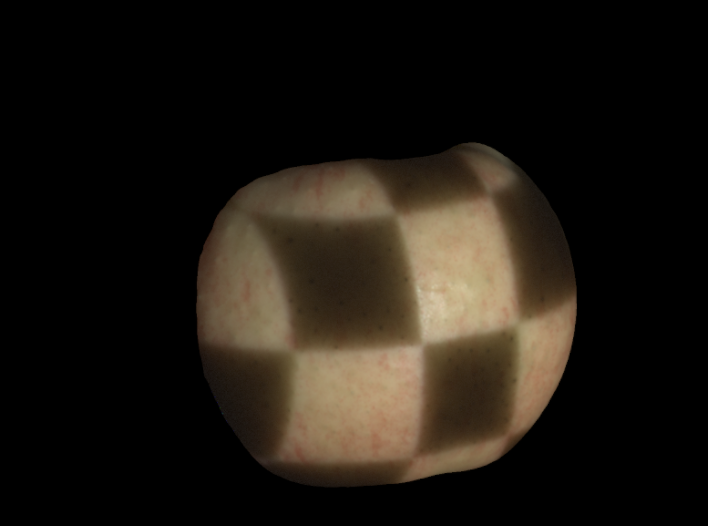} &
\includegraphics[width=0.15\linewidth]{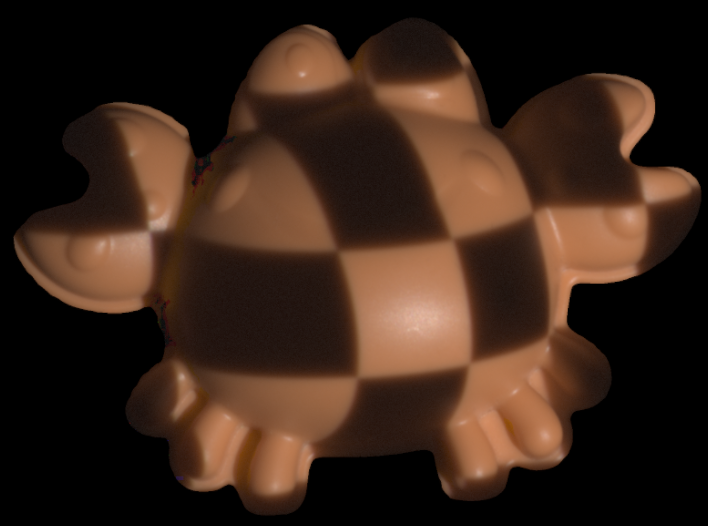} &
\includegraphics[width=0.15\linewidth]{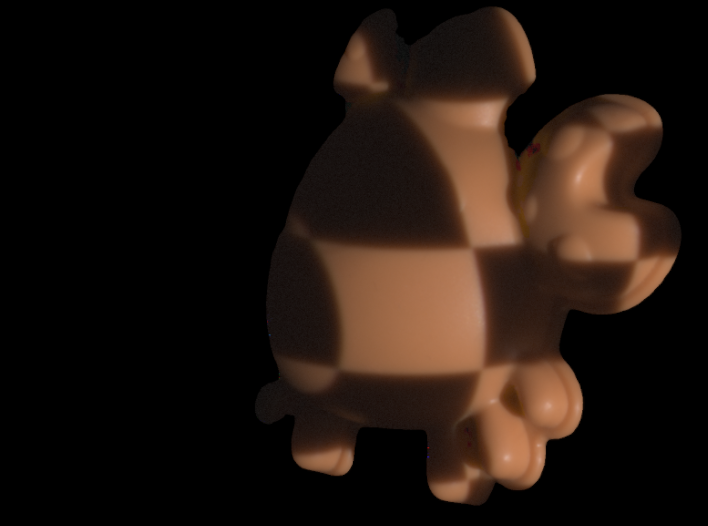} &
\includegraphics[width=0.15\linewidth]{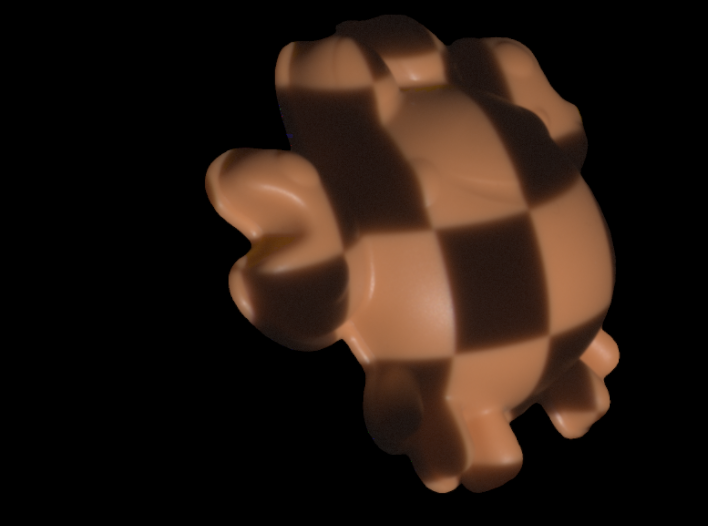} \\

\end{tabular}

\caption{Multiple relit views of two objects using the per-pixel reconstructed SSS footprints compared to camera-captured ground truth, where the real projector illuminates the scene with the same pattern. Note those objects have never been seen during training of the encoder or decoder.}
\label{fig:sss_grid}

\end{figure}

\begin{figure}[htbp]
    \centering
    \setlength{\tabcolsep}{2pt}
    \begin{tabular}{cccc}
        \includegraphics[width=0.23\linewidth]{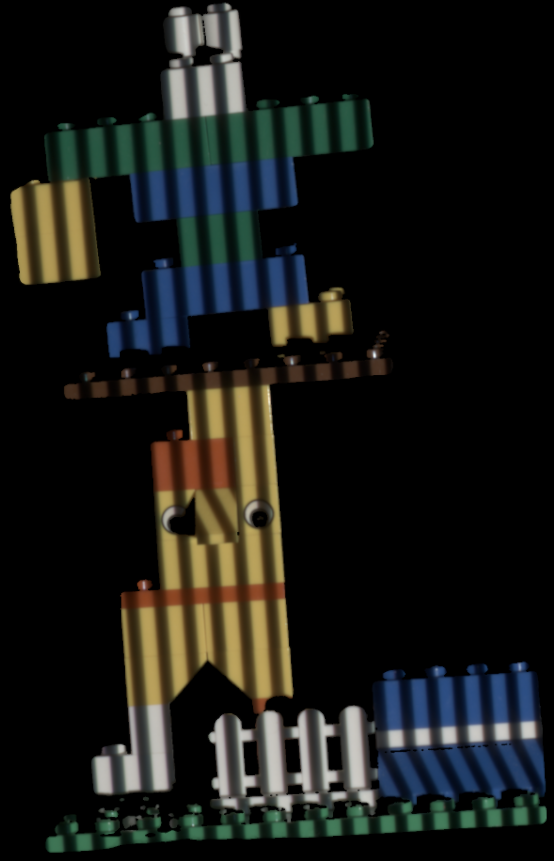} &
        \includegraphics[width=0.23\linewidth]{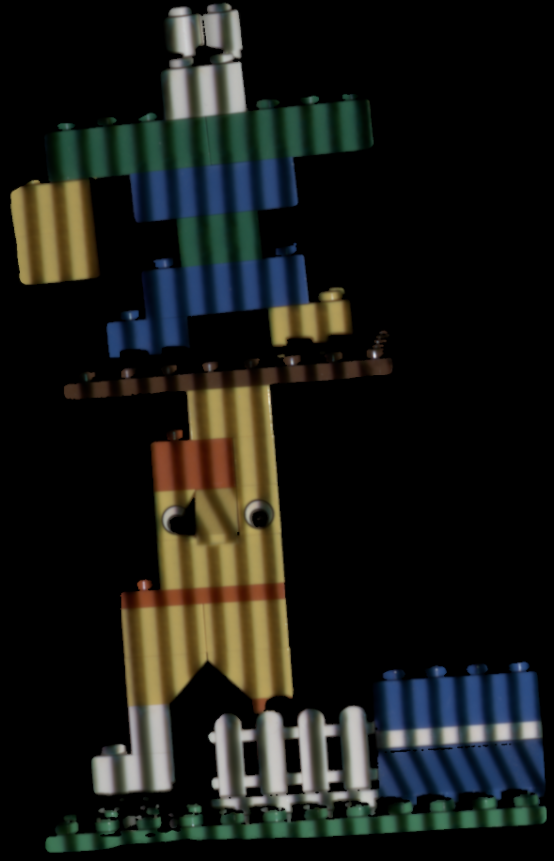} &
        \includegraphics[width=0.305\linewidth]{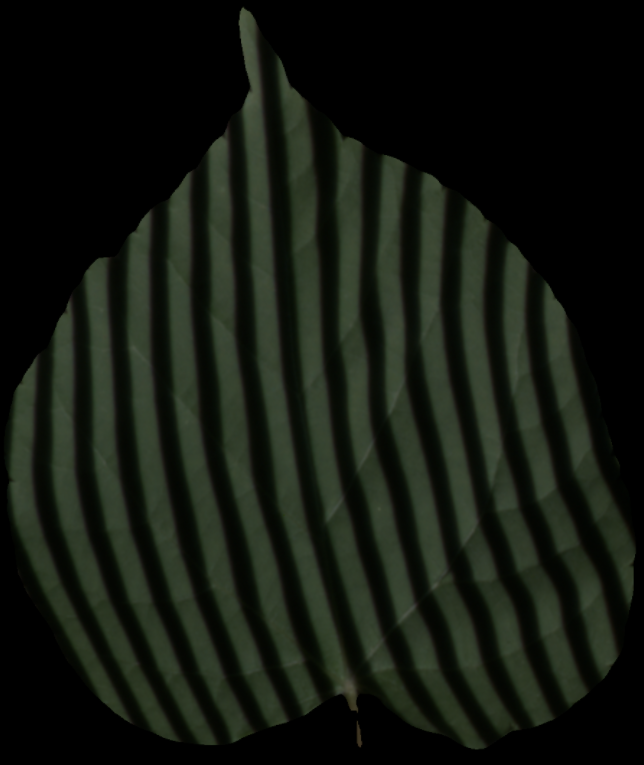} &
        \includegraphics[width=0.305\linewidth]{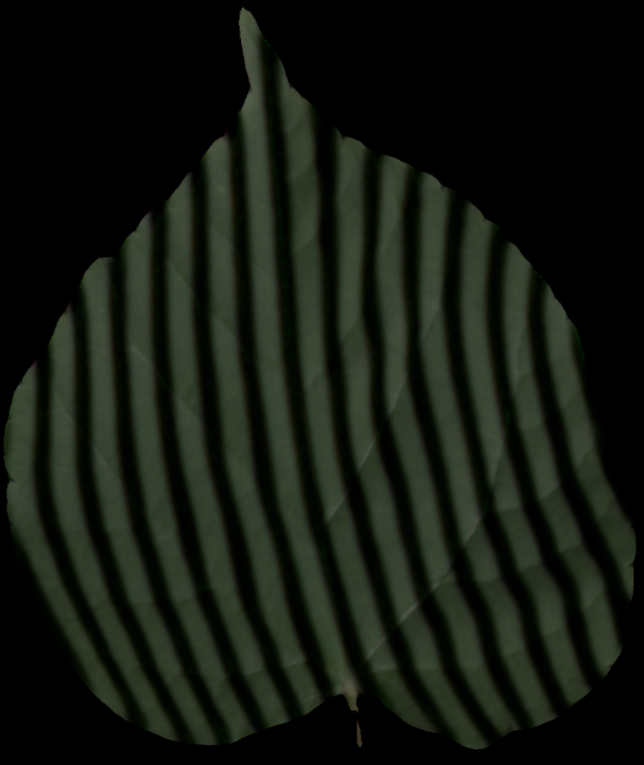} \\
        \small GT & \small Relit & \small GT & \small Relit \\
        \multicolumn{2}{c}{\small LEGO} & \multicolumn{2}{c}{\small Leaf} \\
    \end{tabular}
    \caption{LEGO \& Leaf relit under stripe illumination on unseen objects and illumination.}
    \label{fig:lego_leaf_relit}
\end{figure}

Relighting results for multiple views of the test objects \textit{Apple} and
\textit{Crab} (both unseen during training) are shown in
Fig.\ \ref{fig:sss_grid}. The replicated scattering behavior is of consistent
quality across the views. Besides the accurate surface texture, 
note the subtle color variation in the indirectly lit dark 
checkers of the red and the yellow side of the same apple, 
indicating spatially varying scattering. In the crab example, 
our approach correctly makes some geometric detail visible in 
the dark fields. We demonstrate further generalization by relighting two \textbf{new unseen} objects — a leaf (heterogeneous, fine-veined) and a LEGO brick (anisotropic, edged) — under \textbf{a novel stripe pattern} in Fig.~\ref{fig:lego_leaf_relit}.

Our combined SSL and supervised training scheme reproduces even subtle scattering behavior while drastically reducing the acquisition requirements for any new object: While DISCO \cite{goesele2004disco}, based on individual laser beams, requires millions of images and \cite{ourpaper} requires about 3000 images for novel objects (shifted pixel grids), our approach requires only 8 PSP input images per view for performing accurate inference even on unseen objects. Refer supplemental for more details.

\subsection{kNN Representation Evaluation}

To evaluate the quality of the learned patch embeddings, we perform a $k$-nearest neighbor (kNN) classification experiment using frozen encoder features \cite{unsupervised_feature_learning_instance}. Each object in the dataset is captured from $V$ viewpoints. To avoid data leakage from spatially adjacent patches within the same image, we adopt a \textit{view-level leave-one-out} protocol: all patches from one view are held out for evaluation, while patches from the remaining views form the feature database.

For a query patch, its embedding is compared to the database using cosine similarity, and the label is predicted via majority voting among the $k$ nearest neighbors. We report two complementary metrics: (1) \textbf{overall accuracy}, defined as the fraction of correctly classified patches across all held-out views, and (2) \textbf{mean per-view accuracy}, computed by averaging classification accuracy across views, thereby giving equal weight to each viewpoint.

This evaluation protocol is entirely training-free: the encoder remains frozen
and no fine-tuning or linear probing is performed. Consequently, strong kNN
performance indicates that the learned representations capture discriminative
material characteristics. Quantitative results comparing
PSP-specific augmentations with standard ImageNet augmentations are reported in
Table~\ref{tab:knn_acc_data_aug}. 
consistently achieve higher accuracy. A detailed analysis is in supplemental (6.2).

\subsection{Visualization via PCA-RGB Feature Mapping}
\begin{figure}[t]
\centering
\setlength{\tabcolsep}{2pt}

\begin{tabular}{cccc}
\begin{minipage}[c]{0.24\linewidth}\centering
\rotatebox{90}{\includegraphics[height=3.0cm]{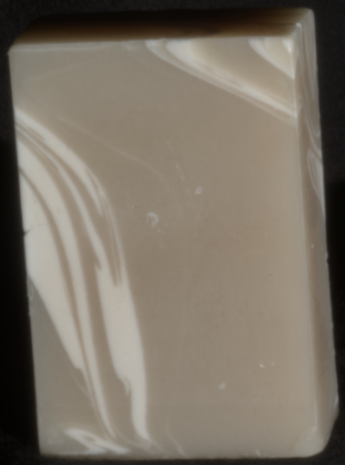}}
\end{minipage}
&
\begin{minipage}[c]{0.24\linewidth}\centering
\rotatebox{90}{\includegraphics[height=3.0cm]{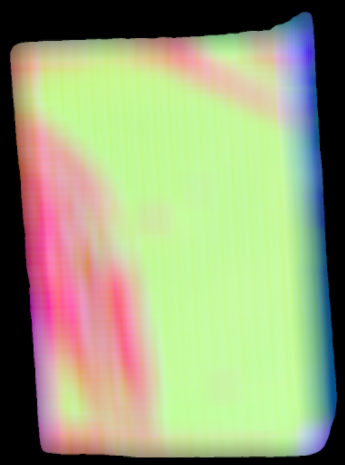}}
\end{minipage}
&
\begin{minipage}[c]{0.24\linewidth}\centering
\rotatebox{90}{\includegraphics[height=3.0cm]{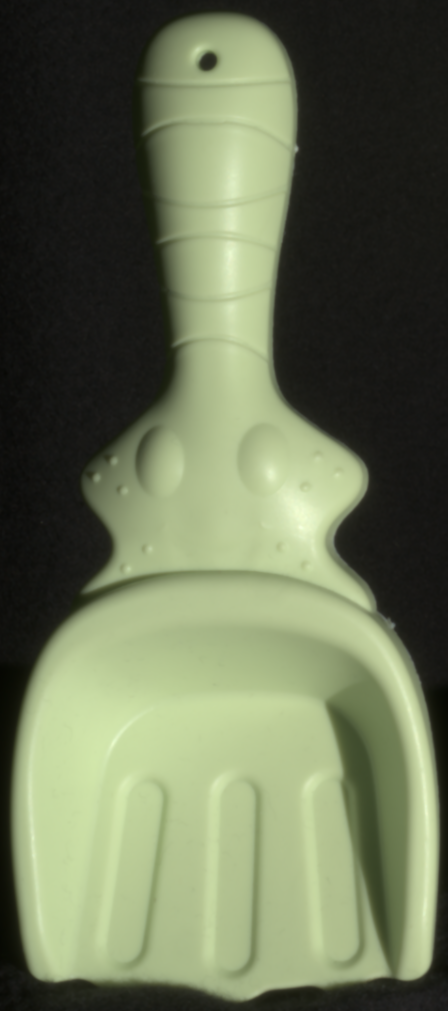}}
\end{minipage}
&
\begin{minipage}[c]{0.24\linewidth}\centering
\rotatebox{90}{\includegraphics[height=3.0cm]{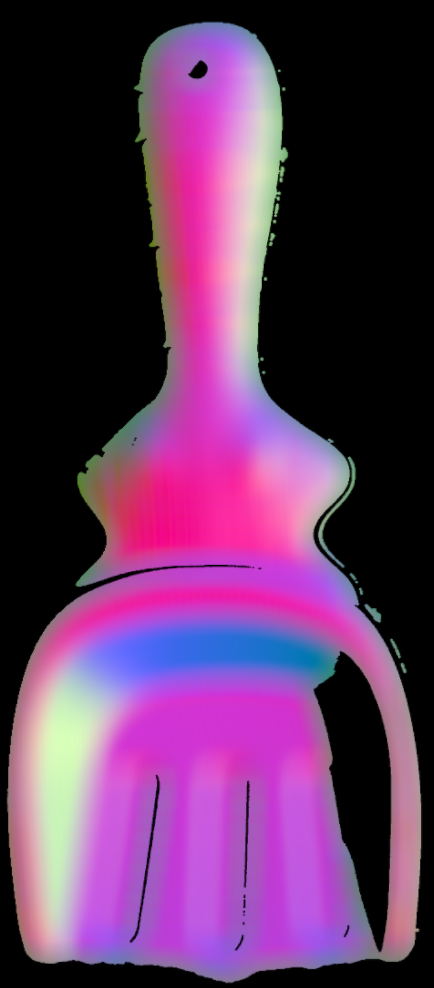}}
\end{minipage}
\\
\multicolumn{2}{c}{Soap [unseen test object]} &
\multicolumn{2}{c}{Hand Shovel [trained on]}
\end{tabular}

\caption{PCA-RGB visualization of the learned SSS features for different objects. On the soap with heterogeneous materials, the visualization clearly marks the veins. On the hand shovel, the normal directions can be seen.}
\label{fig:pca_rgb_soap_shovel}
\end{figure}

To visualize the spatial distribution of SSS properties learned by the
pretrained encoder, we map patch features to colors using a PCA-RGB projection
inspired by~\cite{caron2021dino}. Since the encoder is trained via SimSiam to
produce consistent representations for patches with similar scattering behavior,
spatially similar SSS properties cluster in feature space. PCA preserves this
structure, so similar colors correspond to similar subsurface scattering
characteristics, while color discontinuities indicate material boundaries or
changes in scattering depth. Examples are
shown in Fig.~\ref{fig:pca_rgb_soap_shovel}. For an object with nearly homogeneous
scattering, the visualization resembles surface normal maps due to the
dependence on local illumination directions, whereas spatially varying materials
exhibit clear color variation. Additional examples and details are provided in Supplemental (6.1).

\subsection{Limitations and Future Work}
Our method reconstructs accurate SSS footprints for the specific
camera--projector configuration used during acquisition. Although it generalizes
across views and geometric changes, footprints remain dependent on local
illumination and viewing directions, and interpolation across camera or
projector positions remains future work. Moreover, scattering is restricted to
the local $(90\times90)$ patches, limiting the recoverable scattering radius and
preventing longer-range transport. Finally, reliance on standard dynamic range images may
introduce noise and dynamic-range limitations in the recovered SSS responses.

Future work could explore HDR acquisitions to improve signal quality and
investigate generative or stochastic modeling approaches for SSS footprints
instead of deterministic regression, potentially enabling better representation
of complex or highly anisotropic scattering behavior.

\section{Conclusion}
By using only eight high-frequency PSP images, our model learns to predict accurate, anisotropic scattering footprints that generalize to novel objects and viewpoints. 
This was enabled by splitting the encoder-decoder translation into two separate training tasks: stable self-supervised training of the encoder, which only required easy-to-acquire PSP images to produce a versatile latent representation, followed by supervised training of the decoder to predict accurate scattering impulse-response footprints. 
We showed that domain-specific augmentations and a tailored hybrid loss are critical for maintaining structural and radiometric fidelity. Our method drastically reduces the number of images needed for precise footprint predictions to only 8 images per view. 

\section*{Acknowledgements}
This work has been supported by the Deutsche Forschungsgemeinschaft (DFG) – EXC number 2064/1 – Project number 390727645 and SFB 1233, TP 2, Project number 276693517, and by the International Max Planck Research School for Intelligent Systems (IMPRS-IS).

%
%
\bibliographystyle{splncs04}
\bibliography{main}
\end{document}